# Distributed Real-Time Sentiment Analysis for Big Data Social Streams

Amir Hossein Akhavan Rahnama
Department of Mathematical Information Technology
University of Jyväskylä
Jyväskylä, Finland
amirrahnama@gmail.com

*Abstract*— Big data trend has enforced the data-centric systems to have continuous fast data streams. In recent years, real-time analytics on stream data has formed into a new research field, which aims to answer queries about "what-is-happening-now" with a negligible delay. The real challenge with real-time stream data processing is that it is impossible to store instances of data, and therefore online analytical algorithms are utilized. To perform real-time analytics, pre-processing of data should be performed in a way that only a short summary of stream is stored in main memory. In addition, due to high speed of arrival, average processing time for each instance of data should be in such a way that incoming instances are not lost without being captured. Lastly, the learner needs to provide high analytical accuracy measures. Sentinel is a distributed system written in Java that aims to solve this challenge by enforcing both the processing and learning process to be done in distributed form. Sentinel is built on top of Apache Storm, a distributed computing platform. Sentinel's learner, Vertical Hoeffding Tree, is a parallel decision tree-learning algorithm based on the VFDT, with ability of enabling parallel classification in distributed environments. Sentinel also uses SpaceSaving to keep a summary of the data stream and stores its summary in a synopsis data structure. Application of Sentinel on Twitter Public Stream API is shown and the results are discussed.

*Keywords*— *real-time analytics*, *machine learning, distributed systems, vertical hoeffding tree, distributed data mining systems, sentiment analysis, social media mining, Twitter*

I. INTRODUCTION

In recent years, stream data is generated at an increasing rate. The main sources of stream data are mobile applications, sensor applications, measurements in network monitoring and traffic management, log records or click-streams in web exploring, manufacturing processes, call detail records, email, blogging, twitter posts, Facebook statuses, search queries, finance data, credit card transactions, news, emails, Wikipedia updates [5]. On the other hand, with growing availability of opinion-rich resources such as personal blogs and micro blogging platforms challenges arise as people now use such systems to express their opinions. The knowledge of real-time sentiment analysis of social streams helps to understand what social media users think or express "right now". Application of real-time sentiment analysis of social stream brings a lot of opportunities in data-driven marketing (customer's immediate response to a campaign), prevention of disasters immediately, business disasters such as Toyota's crisis in 2010 or Swine Flu epidemics in 2009 and debates in social media. Real-time sentiment analysis can be applied in almost all domains of business and industry.

Data stream mining is the informational structure extraction as models and patterns from continuous and evolving data streams. Traditional methods of data analysis require the data to be stored and then processed off-line using complex algorithms that make several passes over data. However in principles, data streams are infinite, and data is generated with high rates and therefore it cannot be stored in main memory. Different challenges arise in this context: storage, querying and mining. The latter is mainly related to the computational resources to analyze such volume of data, so it has been widely studied in the literature, which introduces several approaches in order to provide accurate and efficient algorithms [1], [3], [4]. In real-time data stream mining, data streams are processed in an online manner (i.e. real-time processing) so as to guarantee that results are up-to-date and that queries can be answered in real-time with negligible delay [1], [5]. Current solutions and studies in data stream sentiment analysis are limited to perform sentiment analysis in an off-line approach on a sample of stored stream data. While this approach can work in some cases, it is not applicable in the real-time case. In addition, real-time sentiment analysis tools such as MOA [5] and RapidMiner [3] exist, however they are uniprocessor solutions and they cannot be scaled for an efficient usage in a network nor a cluster. Since in big data scenarios, the volume of data rises drastically after some period of analysis, this causes uniprocessor solutions to perform slower over time. As a result, processing time per instance of data becomes higher and instances get lost in a stream. This affects the learning curve and accuracy measures due to less available data for training and can introduce high costs to such solutions. Sentinel relies on distributed architecture and distributed learner's to solve this shortcoming of available solutions for real-time sentiment analysis in social media.

This paper is organized as follows: In section 2, we discuss stream data processing. In section 3, stream data classification is discussed. Section 4 is a discussion on distributed data mining, followed by section 5 about distributed learning algorithms. In section 6, we discuss Sentinel's architecture and lastly, we present the Twitter's public stream case study in section 7 and section 8 includes a brief summary of this paper.

## II. DATA STREAM PROCESSING

Stream data processing problem can be generally described as follows. A sequence of transactions arrives online to be processed utilizing a memory-resident data structure called *synopsis* [1] and an algorithm that dynamically adjusts structure storage to reflect the evolution of transactions. Each transaction is either an insertion of a new data item, a deletion of an existing data item, or any allowed type of query. The synopsis data structure, as well as the algorithm is designed to minimize response time, maximize accuracy and confidence of approximate answers, and minimize time/space needed to maintain the synopsis [4].

Data stream environment has significant differences with batch settings. Therefore each stream data processing method must satisfy the following four requirements in order to be considered:

- Requirement 1: Process an example at a time, and inspect it only once (at most)
- Requirement 2: Use a limited amount of memory
- Requirement 3: Work in a limited amount of time
- Requirement 4: Be ready to predict at any time

The algorithm is passed the next available example from the stream (Requirement 1). The algorithm processes the example, updating its data structures. It does so without exceeding the memory bounds set on it (requirement 2), and as quickly as possible (Requirement 3). The algorithm is ready to accept the next example. On request it is able to predict the class of unseen examples (Requirement 4) [5].

## III. STREAM DATA CLASSIFICATION

In this study, sentiment analysis is formulated as a classification problem. Several predefined categories, which each sentiment can be expressed as, are created. The classifier will decide upon whether a sentiment is expressed in a positive, negative, neutral, … category in an evolving data stream.

*Sequential supervised learning* (i.e. *data stream classification*) problem as follows: Let $\{(x_i, y_i)\}_{i=1}^{N}$ be a set of s $N$ training examples. Each example is a pair of sequences $(x_i, y_i)$, where $x_i =< x_{i,1}, x_{i,2}, ..., x_{i,T_i} >$ and $y_i =< y_{i,1}, y_{i,2}, ..., y_{i,T_i} >$. For example, in part-of-speech tagging, one $(x_i, y_i)$, pair might consist of $x_i = \langle$do you want fries with that$\rangle$ and $y_i = \langle$verb pronoun verb noun prep pronoun$\rangle$. The goal is to construct a classifier $h$ that can correctly predict a new label sequence $y = h(x)$, given an input sequence $x$ [13].

## IV. DISTRIBUTED DATA MINING SYSTEMS

The successful usage of distributed systems in many data mining cases were shown in [9], [10], [12] and [15], Distributed systems increase the performance by forming a cluster of low-end computers and they guarantee reliability by having no single point of failure. Distributed systems are scalable in contrast with monolithic uniprocessor systems. Such features make distributed data mining systems a sound candidate for data stream mining in real-time. In general, distributed data mining systems perform distributed learning algorithms on top of distributed computing platforms in which components are located on networked computers and communicate their actions by passing messages. Distributed systems function according to a topology. *Topology* is a collection of connected processing items and streams. It represents a network of components that process incoming data streams. A *processor* is a unit of computation element that executes parts of algorithm on a specific Stream Processing Engine (SPE). Processors contain the logic of the algorithms. *Processing items* (PI) are the internal different concrete implementation of processors.

Distributed Systems pass content events through streams. A *stream* is an unbounded sequence of tuples. A tuple is a list of values and each value can be any type as long as the values and each value can be any type as long as the values are serializable, i.e. dynamically typed. In other words, a stream is a connection between a PI into its corresponding destinations PIs. Stream can be seen as a connector between PIs and mediums to send content events between PIs. A content event wraps the data transmitted from a PI to another via a stream. A source PI is called a spout. A *spout* sends content events through stream and read from external sources. Each stream has one spout. A *bolt* is a consumer of one of more input streams. Bolts are able to perform several functions for the input stream such as filtering of tuples, aggregation of tuples, joining multiple streams, and communication with external entities (caches or database). Bolts need *grouping* mechanisms, which determines how the stream routes the content events. In *shuffle* grouping, stream routes the content events in a round-robin way to corresponding runtime PIs, meaning that each runtime PI is assigned with the same number of content events from the stream. In all grouping, stream replicates the content events and routes them to all corresponding runtime PIs. In *key* grouping, the stream routes the content event based on the key of the content event, meaning that content events with the same value of key are always routed into the same runtime PI.

---

[1] discussed in more details in section 6.

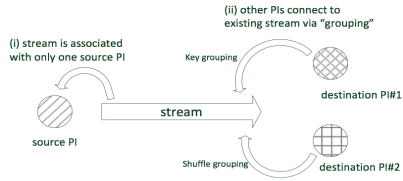

Fig 1. Instatiation of a Stream and Examples of Groupings [2]

Transformation of stream between spouts and bolts follows the *pull* model, i.e. each bolt pulls tuples from the source components (which can be bolts or spouts). This implies that loss of tuples happen in spouts when they are unable to keep up with external event rates.

There are two types of nodes in a Storm cluster: master node and worker node. *Master node* runs the Nimbus daemon that is responsible for assigning tasks to the worker nodes. A nimbus is the master node that acts as entry point to submit topologies for execution on cluster. It distributes the code around the cluster via supervisors. A supervisor runs on slave node and it coordinates with ZooKeeper to manage the workers. A *worker* corresponds to JVM process that executes a part of topology and comprises of several executors and tasks. An executor corresponds to a thread spawned by a worker and it consists of one or more tasks from the same bolt or spouts. A task performs the actual data processing based on spout/bolt implementation.

## V. DISTRIBUTED LEARNING ALGORITHMS

Parallelism type refers to the ways a distributed learning algorithm performs its parallelization. There are three types of parallelism: horizontal, vertical and task parallelism. In order to go deeper in each type of parallelism, we need to define a few concepts. A *processing item* (PI) is a unit of computation element (a node, a thread of process) that executes some part of algorithm. A *user* is a client (human being or a software component such as a machine learning framework) who executes the algorithm. *Instance* is defined as a datum in the training data. Meaning that, the training data consists of a set of instances that arrive once at a time to the algorithm. In *horizontal parallelism*, SPI sends instances into distributed algorithm. The advantage of horizontal parallelism lies in the fact that it is suited for very high arrival rates of instances. The algorithm is also flexible in the fact that it allows the user to add more processing power. In *vertical parallelism*, The difference is that local-statistic PIs do not have the local model as in horizontal parallelism and each of them only stores sufficient statistic of several attributes that are assigned to it and computes the information-theoretic criteria based on that assigned statistic. *Task parallelism* consists of sorter processing item and updater-splitter processing item, which distributes model into available processing items.

## VI. SENTINEL: ARCHITECTURE & COMPONENTS

### A. Programming model

While map-reduce is the most popular programming model for big data scenarios, map-reduce model is inapplicable to data stream processing. Map-reduce operations are not I/O efficient, since map and reduce are blocking operations, therefore a transition to the next stage cannot be done until all tasks of the current stage are finished. Consequently, pipeline parallelism cannot be achieved. One key drawback is the poor performance of map-reduce due to the fact that all the input should be already prepared for a map-reduce job in advance which causes a high latency in working with map-reduce algorithms [3].

Sentinel runs on top of Apache Storm. *Apache Storm*[2] is a free and open source distributed real-time computation system. Storm makes it easy to reliably process unbounded streams of data. Storm has many use cases: real-time analytics, online machine learning, continuous computation, distributed Remote Procedure Calls (RPC), Extract Transform Load (ETL), and more. Storm allows the computation to happen in parallel in different nodes, which can be in different clusters. This feature enables parallel pipeline of data, which makes Storm a perfect framework for data stream mining settings.

### B. Overall Architecture

In Sentinel, input social stream is read from the source and instances of stream continuously are read by ADWIN with an adaptive window. ADWIN node reads the data and checks the source distribution along with arrival rate of instances and adapts the window size to the speed and volume of incoming instances. Then it passes the instances to the data pipeline node. In data pipeline node, first adaptive filtering component filters the instances based on the desired attribute and converts the data into a vector format and passes it onwards. Feature selector summaries the text from the incoming instances[3] and passes it to the frequent item miner algorithm. Frequent item miner algorithms keep a summary of the text tokens and number of appearances in the document. The resulting hash table will be ready to be sent to the Vertical Hoeffding Tree learner's node. Source Processing Item(s) (SPI) take the input from the passed hash table and passes it to the model and their local statistic APIs. The evaluator-processing item updates the result of learning onto the synopsis. The summary of data can be stored in the archive database in specific long time intervals per day. This is an online real-time process and the raw instance is never saved in any node or components. On the other end of the system, the user will be able to query the synopsis at anytime and the result will be the coming from the model.

---

[2] http://storm.incubator.apache.org/
[3] In social stream, instance are in form of documentss

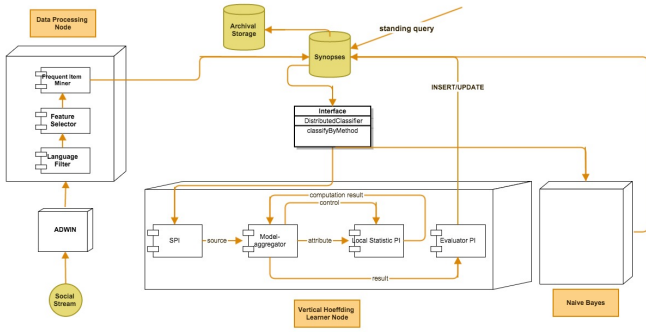

Fig 2. Sentinel Architecture

## C. ADWIN

In proposed solution, we use sliding window models in conjunction with learning and mining algorithms, namely modified version of ADWIN[4] (Adaptive Windowing) algorithm that maintains a window of variable size. ADWIN grows the size of sliding window in case of no change and shrinks it when changes appear in the stream [7].

## D. Synopses

Synopsis data structure is substantively smaller than their base data, resides in main memory, transmitted remotely at minimal cost and transmitted remotely at minimal cost [14].

## E. Data Pipeline Node

### 1) Frequency and Language Filter

In this component, after that the tweets are filtered by their language[5], they are converted to sparse vectors by *tf-idf* filter. This is to make the instances ready for feature selection.

$$tf - idf_{t,d} = tf_{t,d} \times idf_t$$

Where $tf_{t,d}$ is the frequency of term *t* in document *d*. . Inverse document frequency, $idf_t$ is as follows:

$$idf(t, D) = \log \frac{N}{|\{d \in D : t \in d\}|}$$

is a logarithm of *N* as total number of documents in the corpus divided by total number of documents containing the term *d*.

### 2) Feature selector

---

[4] called *ADWIN2* which has improvements in performance over its older version, *ADWIN*
[5] language filter is based on a Naive Bayesian Classifier from language detection library with a guarantee of precision up to 99%, available at https://code.google.com/p/language-detection/

Data reduction techniques are used to get a smaller volume of the data. Sentinel considers each document as a list of words. Adaptive filter will transform them to vectors of features, obtaining the most relevant ones. We use an incremental *tf-idf* weighting scheme:

$$f_{i,j} = \frac{freq_{i,j}}{\sum_l freq_{l,j}}$$

$$idf_i = \log \frac{N}{n_i}$$

where $f_{i,j}$ is the frequency of term i in document j that is the number of occurrences of term i in document j divided by the sum of frequency of all terms in document j, i.e. the size of the document. $idf_i$ is the inverse document frequency of term i. *N* is the number of documents and $n_i$. This approach improves the performance of using synopsis by keeping only the most relevant words within a document into the synopsis data structure.

### 3) Frequnet Item Miner

Several algorithms have been proposed to enable balance between the infinity of data streams compared to finite storage capacity of a computing machine. The core idea behind such algorithms is that only a portion of the stream gets to be stored. Frequent item miners have three different categories: *Sampling-based, Counting-based algorithms* and *Hashing-based algorithms*. For our purpose, we store the tokens with their number of appearances in the stream. Therefore, the frequent item miner algorithm will need to be from counting-based category. In this study, based on the extraordinary performance result of measures in precision and recall of different Count-based algorithms in [4], *Space Saving* algorithm was chosen due to having recall and precision close to 92%.

*Space Saving* was proposed for hot-list queries under some assumptions on the distribution of the input stream data. The space-saving algorithm is designed to estimate the frequencies of significant elements and store these frequencies in an always-sorted structure, accurately. The gain in using *Space-Saving* in our proposed solution is that it returns not only ε-deficient frequent items for queries, but also guarantees and sorts top-*k* items for hot-list queries under appropriate assumptions on the distribution of the input.

## F. Vertical Hoeffding Tree Node

Vertical Hoeffding Tree [2] is our selection for distributed stream mining classifier. VHT is based on VFDT (Very Fast Decision Tree) with vertical parallelization. In social stream settings, stream mining algorithm is applied to instances in form of documents. Social streams involve instances with high number of attributes therefore VHT is a suitable candidate due to its vertical parallelism approach. VHT's vertical parallelism brings advantages over other types of parallelism types in this context. Also, since the learning algorithm is based on VFDT, it applies well to social streams with high speed of arrival of data instances.

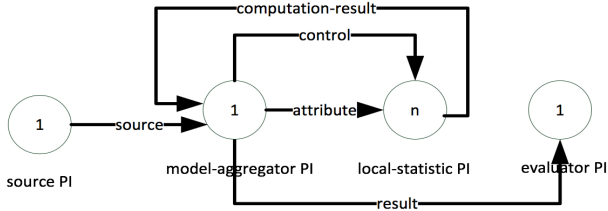

Fig. 3: Vertical Hoeffding Tree: Components & Process [2]

In VHT algorithm, each node has a corresponding number that represents its level of parallelism. Model-aggregator PI consists of the decision tree model and connects to local-statistic PIs through *attribute* and *control* stream. Model aggregator splits instances based on attribute and each local-statistic PI contains local statistic for such attributes that was assigned to them. Model-aggregator PI sends the split instances through attribute stream and it sends control messages to ask local-statistic PI to ask local-statistic PIs to perform computation via control stream.

Model-aggregator receives *instance* content events from source PI and it extracts the instances from content events. After that, model-aggregator PI needs to split the instance based on the attribute and send *attribute* stream to update the sufficient statistic for the corresponding leaf. When a local-statistic PI receives attribute content event, updates its corresponding local statistic. To perform this, it keeps a data structure that store local statistic based on leaf ID and attribute ID. When it is the time to grow the tree, model-aggregator sends compute content event via *control* stream to local-statistic PI. Upon receiving compute content event, each local-statistic PI calculates $\bar{G}_l(X_i)$ [6] to determine the best and second best attributes. The next part of the algorithm is to update the model once it receives all computation results from local statistics.

Whenever the algorithm receives a local-result content event, it retrieves the correct leaf *l* from the list of the splitting leaves. Then, it updates the current best attribute $X_a$ and second best attributes $X_b$. If all local results have arrived into model-aggregators PI, the algorithm computes Hoeffding bound and decides whether to split the leaf *l* or not. To handle stragglers, model-aggregator PI has time-out mechanism to wait for computation results. If the time out happens, the algorithm uses current $X_a$ and $X_b$ to compute Hoeffding bound and make the decision[7].

## VII. CASE STUDY: TWITTER PUBLIC STREAM API

Twitter currently provides a Stream API and two discrete REST APIs. Through the stream API users can obtain real-time access to tweets. Twitter public Stream API[8] is the example of social stream that we showcase Sentinel. Common problem in unbalanced data streams such as Twitter is that classifiers have high accuracy, close to 90% due to the fact that a large portion of falls into one of the classification classes. This is more apparent in Twitter Stream however, as mentioned before, in projects such as Sentiment 140[9], data does not constitute a representative sample of the real Twitter stream due to the fact that the data is pre-processes, balanced and has shrunk in size to obtain a balanced and representative sample.

### A. Training

In this study, we converted the raw data into a new vector format as in [6]. Data pipeline particularly for this case study, performs the feature reduction and labeling via emoticons during the model's testing phase as follows:

- *Feature Reduction*: Data pipeline replaces words starting with the @ symbol with the token *USER*, and URLs within the same Tweet by the token URL.
- *Emoticons*: Data pipeline uses emoticons to generate class labels during the training phase of the classifier however after that all emoticons are deleted.

To measure accuracy and performance of learning algorithm, a forgetting mechanism with a sliding window of most recent observation can be used [6]. It was shown that prequential evaluation is not a reliable measure for unbalanced data streams and proposed Kappa as an evaluation measure. In this study, we show that based on a sliding window and with usage of Kappa statistic this issue is solved [5].

### B. Experiment

For the experiment, we ran sentinel on a three-node cluster. Each node had 48GB of RAM and quad core Intel Xeon 2.90GHZ CPU with 8 processors. We ran the experiment with a sample of 1 million tweet instances. We have filtered the tweets to only English tweets tree learning algorithms, which were used, were Multinomial Naïve Bayes, Hoeffding Tree, and Vertical Hoeffding Tree. In case we have followed an offline approach for each million tweets, 1GB of disk space was needed, however since we follow an online approach, there is no need for disks. Due to the release of iPhone 6 to the date of this publication, we focused on performing a sentiment analysis on the newest iPhone. We trained our three learners with query "iPad" and we tested the model with query "iOS 8". It should be noted that generally in Twitter or most social networks, users have more positive or almost positive sentiment rather than negative ones.

---

[6] G is information gain measure however it can be replaced with other statistic measures such as Gini index or gain ratio.
[7] See [2] for a complete pheducecode of different steps of the algorithm.

[8] https://dev.twitter.com/docs/streaming-apis/streams/public
[9] http://www.sentiment140.com/

TABLE I. MEASURES OF DIFFERENT CLASSIFIERS IN SENTINEL

| Classifiers | Accuracy Measures & Processing Time | |
|---|---|---|
| | Kappa | Time |
| Multinomial Naïve Bayes | 57.78% | 3123 sec. |
| Hoeffding Tree | 66.20% | 4017 sec. |
| Vertical Hoffding Tree | 78.57% | 1309 sec. |

a.

As you can see in Table 1, Vertical Hoeffding Tree performs significantly better both in accuracy and in time compared to the other classifiers. Multinomial Naïve Bayes classifier is faster than Hoeffding Tree however it is less accurate. One of the main reasons is that VHT is based on VFDT Figure 4 show the learning curve of classifiers with sliding window of 10000 instances per window. It should be mentioned that due to memory and speed limitation in labeling stream instances in online approaches, the learners have less accurate results compared to online approaches.

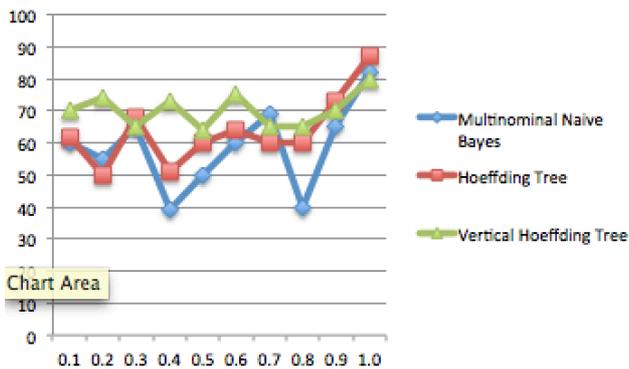

Fig 4. Sliding window Kappa Statistic (%) per millions of instance

## VIII. CONCLUSION

In this study, we presented a distributed system to perform real-time sentiment analysis. After discussing data stream mining and distributed data mining systems, different components of the system were discussed. The learning algorithm of the solution is based on Vertical Hoeffding Tree, a parallel decision tree classifier. We ran the solution against Multinomial Naïve Bayes and Hoeffding Tree classifiers and compared the results which showed significant both accuracy and performance improvement compared to uniprocessor classifiers.